\documentclass[letterpaper, 10 pt, conference]{ieeeconf}
\IEEEoverridecommandlockouts    % This command is only needed if
\usepackage{graphics}           
\usepackage{times}              
\usepackage{amsmath}            
\usepackage{amssymb}            
\usepackage{graphicx}
\usepackage{algorithm}
\usepackage[noend]{algpseudocode}
\usepackage{booktabs}
\usepackage{color}
\usepackage{listings}
\usepackage{subfiles}
\usepackage{hyperref}
\usepackage[nocompress]{cite} % sorts citations automatically, e.g., "[1],[2],[3]" instead of "[2],[3],[1]".
\definecolor{instructioncolor}{rgb}{.5,.5,.5}

\usepackage[font=small]{caption}

\def\secref#1{Sec.~\ref{#1}}
\def\figref#1{Fig.~\ref{#1}}
\def\tabref#1{Tab.~\ref{#1}}
\def\eqref#1{Eq.~(\ref{#1})}

\makeatletter
\usepackage{xspace}
\DeclareRobustCommand\onedot{\futurelet\@let@token\@onedot}
\def\@onedot{\ifx\@let@token.\else.\null\fi\xspace}

\def\etal{{et al}\onedot}
\makeatother

\usepackage{array}
\newcolumntype{L}[1]{>{\raggedright\let\newline\\\arraybackslash\hspace{0pt}}m{#1}}
\newcolumntype{C}[1]{>{\centering\let\newline\\\arraybackslash\hspace{0pt}}m{#1}}
\newcolumntype{R}[1]{>{\raggedleft\let\newline\\\arraybackslash\hspace{0pt}}m{#1}}

\def\argmax{\mathop{\rm argmax}}

\def\argmin{\mathop{\rm argmin}}

\newcommand{\RR}{\mathbb{R}}

\newcommand{\norm}[1]{\lVert#1\lVert}

\renewcommand{\b}[1]{\mbox{\boldmath$#1$}}

\renewcommand{\v}[1]{{\b #1}} 

\newcommand{\m}[1]{{\mbox{{\sffamily\slshape{#1\/}}}}}

\usepackage{cite}
\usepackage{transparent}
\usepackage{flushend}
\makeatletter
\newcommand{\linebreakand}{%
  \end{@IEEEauthorhalign}
  \hfill\mbox{}\par
  \mbox{}\hfill\begin{@IEEEauthorhalign}
}
\makeatother

\title{\LARGE \bf Sparse Anchors for Visual Multi-Session Registration\\ in Changing Environments}
\author{Yunus Talha Erzurumlu \and Niklas Trekel \and Saurabh Gupta \and Nico Koltermann \and Olga Vysotska \linebreakand Jens Behley \and Louis Wiesmann \and  Cyrill Stachniss% <-this % stops a space
 \thanks{All authors except Olga Vysotska are with the University of Bonn, Center for Robotics. Olga Vysotska is with ETH Zurich and Cyrill Stachniss is additionally with the Lamarr Institute for Machine Learning and Artificial Intelligence.}%
\thanks{This work has partially been funded
by the Deutsche Forschungsgemeinschaft (DFG, German Research Foundation)
under Germany's Excellence Strategy, EXC-2070 -- 390732324 -- PhenoRob,
by the German Federal Ministry of Research, Technology and Space~(BMFTR)
under the Robotics Institute Germany~(RIG),
and by Innosuisse -- Swiss Innovation Agency.
}%}%
}

\begin{document}
\thispagestyle{empty}
\pagestyle{empty}
\maketitle

%%%%%%%%%%%%%%%%%%%%%%%%%%%%%%%%%%%%%%%%%%%%%%%%%%%%%%%%%%%%%%%%%%%%%%%%%%%%%%%%
\begin{abstract}
Repeated robotic monitoring requires localization across visits to the
same environment, even if the environment changes.
Such changes can substantially reduce persistent visual content, and 
repetitive structures can lead to incorrect cross-session correspondences.
In this paper, we address visual multi-session registration of a new data
recording (the query session) to a previously reconstructed reference map.
Under substantial visual and structural changes, cross-session
localization may be possible only for parts of the trajectory.
In our approach, we therefore rely on a sparse set of reliable absolute pose constraints and
combine them with within-session relative-pose information to register the
complete trajectory to the reference map.
We first estimate absolute alignment hypotheses through visual place recognition,
local feature matching, and geometric verification.
We then use sequence consensus to reject hypotheses that are inconsistent with
a common query-to-reference alignment and retain the remaining poses as
absolute anchors.
These anchors are refined through featuremetric alignment against a Gaussian
feature field and combined with within-session relative-pose constraints in a
robust pose-graph optimization.
We evaluate our approach on 15 cross-session experiments from the challenging
HILTI26 construction-site dataset, covering substantial scene changes and
difficult-to-match environments.
Our method achieves 94.4\% recall within 1\,m and 100.0\% within 5\,m,
outperforming independent image localization, which peaks at 65.3\% and
70.7\%, respectively, as well as direct registration of independently
reconstructed Structure-from-Motion geometry.
Our results show that sparse, reliable absolute anchors can provide accurate
and complete trajectory registration even when reliable cross-session
correspondences are available only for part of the query session.
\vspace{-0.2cm}
\end{abstract}

  %
  %% WHY 
  % Use 1-2 not too long sentences, which clearly answer the WHY question: 
  % Why is this relevant, why should I care? Motivate why the stuff that 
  % you enable is relevant (not neccessarily equal to the technique)

  %% WHICH PROBLEM 
  % One sentence that explain the problem the paper addresses/ivestigates
  % Start with: In this paper, we address the problems of \dots

  % In this paper, we  ...

  %% HOW & WHAT
  % Around 3 sentences that explain how to approach the problem in general and answers:
  % How to solve the problem in general? (1/2 - 1 sentence)
  % What makes our approach special? What are we actually doing? What is new?
  
  %% IMPLEMENTATION, EVALUATION, WHAT FOLLOWS
  % 1-2 sentences what the experiments show and potentiall what follows from
  % your great work for the research community or the rest of the world ;-)

%%%%%%%%%%%%%%%%%%%%%%%%%%%%%%%%%%%%%%%%%%%%%%%%%%%%%%%%%%%%%%%%%%%%%%%%%%%%%%%%
\section{Introduction}
\label{sec:intro}

%% --- WHY ---

Most robotic systems operating repeatedly in the same environment must reliably
localize in a previously built map despite scene changes.
Reliable cross-session localization and registration are therefore important for
long-term autonomy, repeated inspection and monitoring, as well as persistent
mapping~\cite{kim2013tro,wang2020icra,hutter2018fsr}.
This becomes particularly challenging when changes are structural rather than purely
appearance-level, since parts of the previously mapped geometry may no longer exist,
while new structures may alter or occlude previously visible content.

Construction sites are a particularly demanding instance of this problem.
Their geometry and appearance evolve as walls, installations, temporary
structures, and surface finishes are added, removed, or modified, while
repetitive and low-texture structures create additional perceptual
ambiguity~\cite{sun2025isprs}, further compounded by the frequent lack of GNSS
availability in indoor construction environments.
% Indoor construction environments are further challenging because GNSS is often
% unavailable.
Construction progress monitoring requires repeated observations to compare the current
as-built state against earlier observations or the planned state
~\cite{tuttas2014isprs}.
Autonomous mobile robots can acquire such observations, but comparing visits requires
reliable localization in a common coordinate frame.
\begin{figure}[t]
  \centering
  \includegraphics[width=0.94\linewidth]{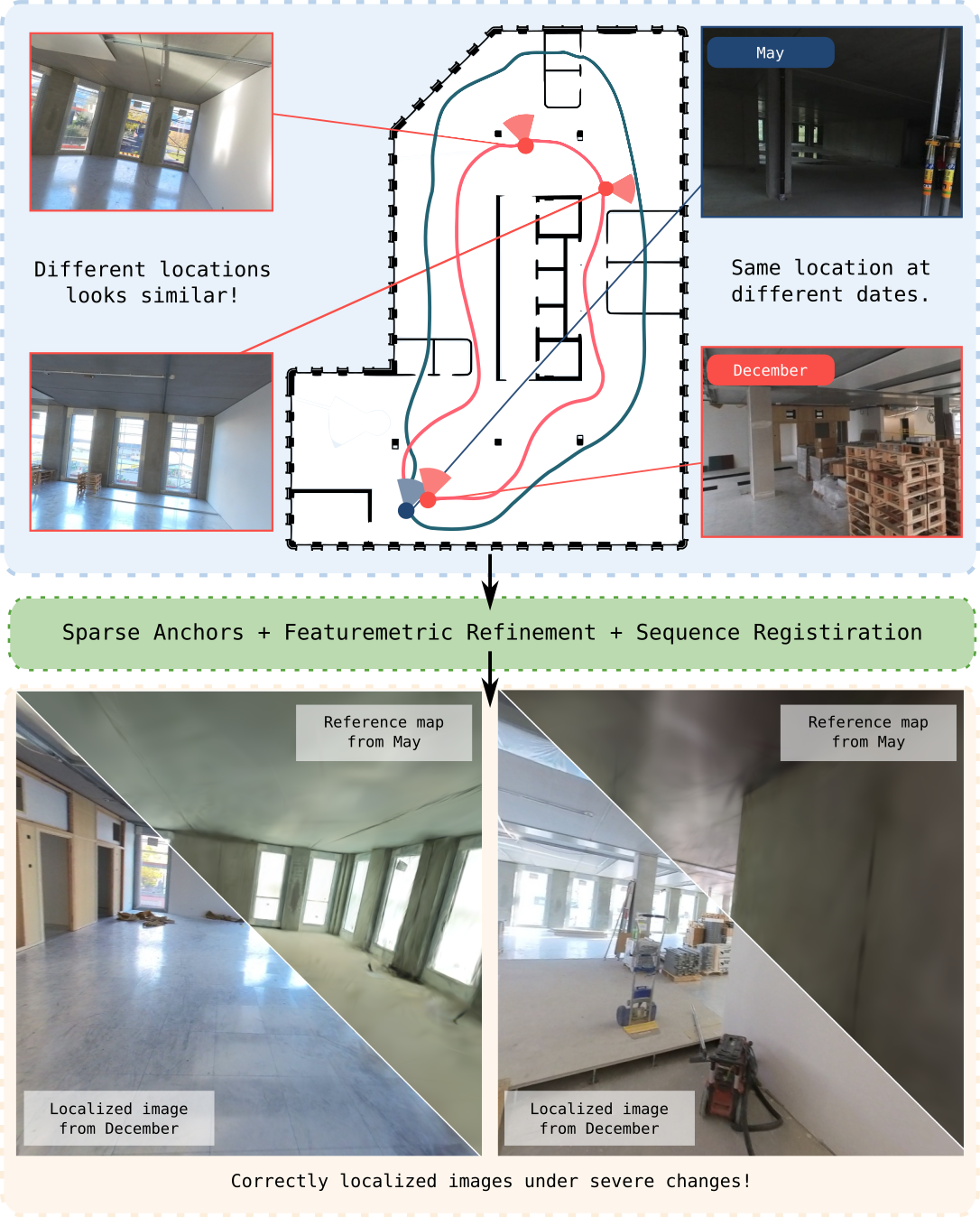}
  \caption{%
    Repetitive structures create perceptual aliasing between distant locations, while
    construction progress causes substantial appearance and geometric changes over time
    (top).
    Using sparse absolute anchors, featuremetric refinement, and sequence-level
    registration, our approach recovers poses with accurate image-to-map alignment even
    under severe scene changes (bottom).
  }
  \label{fig:motivation}
  \vspace{-0.05cm}
\end{figure}

%% --- WHICH PROBLEM ---

In this work, we address visual multi-session registration in challenging,
GNSS-denied, and changing environments.
Given a previously reconstructed \emph{reference map}, together with
its posed reference observations, for example computed with any SLAM system, 
we seek to register the \emph{query session}, a visual sequence acquired at a different time.
The sequence of body poses associated with the noisy query observations forms the
\emph{query trajectory}, which we seek to recover in the reference map coordinate frame.
This is difficult because substantial geometric and appearance changes reduce the
persistent visual content available for cross-session correspondences, while repetitive
structures can produce plausible correspondences between physically distinct locations,
as shown in \figref{fig:motivation}.

%% --- HOW & WHAT (Prior Work & Challenges) ---

Existing strategies often use image-based localization against an earlier
map~\cite{sarlin2019cvpr} or direct registration of independently reconstructed 3D maps.
Image-based localization becomes unreliable when persistent visual content is limited
or perceptually ambiguous, while 3D point-cloud registration degrades as spatial and
temporal changes accumulate~\cite{sun2025isprs}.

%% --- OUR APPROACH / KEY IDEA ---

% To handle scene change and perceptual aliasing, we seek reliable absolute poses only for
% a subset of query observations rather than requiring cross-session correspondences along
% the entire trajectory.
% We perform visual place recognition, distinctiveness filtering, local feature matching,
% and geometric verification to obtain absolute pose hypotheses.
% We retain only those hypotheses that are jointly consistent with a common alignment
% between the query trajectory and the reference map, forming a sparse set of
% \emph{absolute anchors}.
% We refine these anchors by aligning query image features against the reference feature
% field, improving image-to-map alignment for downstream tasks such as change detection
% and construction progress monitoring.
% Finally, we combine the refined anchors with within-session relative-pose constraints
% in a robust pose-graph optimization (PGO) to recover the complete query trajectory in the
% reference map frame. %% coordinate frame is deleted. not sure about this one. 

%% --- MAIN CONTRIBUTION & CLAIMS ---

The main contribution of this paper is a visual multi-session registration approach for
changing environments, where scene changes and perceptual aliasing can make reliable
cross-session correspondences available only for parts of the trajectory.
Our approach combines conservative cross-session anchor selection, retaining only
absolute pose hypotheses that are geometrically verified and consistent with the query
trajectory, with featuremetric pose refinement for precise image-to-map alignment.
These sparse \emph{absolute anchors} are then combined with within-session relative-pose
constraints to recover the complete query trajectory even when reliable cross-session
correspondences are only sparsely available.

We make three key claims: our approach
(i)~provides more accurate and complete multi-session registration than the baselines;
(ii)~improves the reliability of cross-session constraints under scene change and
perceptual aliasing through conservative filtering and sequence consensus; and
(iii)~improves fine image-to-map alignment through featuremetric refinement of the
verified absolute anchors.
We will publicly release our implementation upon publication.
\section{Related Work}
\label{sec:related}

% Discuss the main related work and cite around key papers. The number
% of cites depend on the page limit. Never below 15, 20-30 is often fine
% for conference papers.
% The related work section should be approx. 1 column long, assuming 
% a 6-page paper, up to one page for 8-10 page papers.  You can structure 
% the  section in paragraphs, grouping the  papers, and describing the key 
% approaches with 1-2 sentences. Avoid dull enumerations of papers. If 
% applicable, describe the key difference to your approach at the end 
% of each paragraph briefly. Avoid adding subsections, al least for a 
% conference paper.

A common approach to visual multi-session registration is to localize observations
from a later session against a previously reconstructed reference map.
Visual place recognition~(VPR) identifies corresponding places across sessions
\cite{keetha2024ral,berton2024ral,li2025ral,hausler2025ral}.
Modern VPR methods range from learned aggregation approaches such as
NetVLAD~\cite{arandjelovic2016cvpr} to scalable and viewpoint-robust descriptors
\cite{berton2022cvpr,alibey2023wacv} and representations based on large
pretrained visual features
\cite{keetha2024ral,izquierdo2024cvpr,berton2025cvprws}.
Hierarchical Localization~(HLoc) by Sarlin \etal~\cite{sarlin2019cvpr}
combines retrieval, local feature matching, and perspective-n-point
(PnP)-based localization and forms our image-based localization baseline.
However, such approaches rely on distinctive, persistent cross-session
correspondences and often degrade under appearance changes and repetitive
structures~\cite{sattler2018cvpr,torii2015tpami}.
In contrast, we use image-based localization to obtain sparse, reliable
absolute anchors rather than localizing every query observation.

Leveraging multiple observations can improve robustness when individual cross-session
correspondences are unreliable.
Multi-sequence maps retain observations acquired under different appearance
conditions~\cite{vysotska2019ral}, while sequence-based VPR exploits temporal
context to resolve ambiguities difficult from individual images
\cite{berton2024ral,li2025ral,vysotska2025icra}.
Stenborg \etal~\cite{stenborg2020threedv} combine map-based localization with
relative motion to bridge observations that cannot be localized independently,
but assume that the query approximately follows the reference route.
More generally, multi-map and multi-session SLAM systems exploit cross-session
constraints across different traversals~\cite{campos2021tro}, and HILTI23
benchmarks multi-session SLAM in construction environments~\cite{nair2024ral}.
Our approach similarly exploits within-session motion, but uses sparse
cross-session anchors to register trajectories without requiring the query path
to follow the reference trajectory.

A different strategy is to reconstruct each session independently and register
the resulting 3D maps.
Global point-cloud registration methods address limited overlap and high outlier
rates~\cite{yang2021tro,lim2025icra}, but most target dense LiDAR or RGB-D
geometry rather than sparse and non-uniform Structure-from-Motion~(SfM)
reconstructions.
ColabSfM~\cite{edstedt2025cvpr} specifically addresses this domain gap, while
construction-site studies show that temporal scene change further reduces
persistent geometric overlap~\cite{sun2025isprs}.
Rather than aligning independently reconstructed geometry, our method
uses sparse image-to-map pose constraints together with the query trajectory.

Featuremetric localization refines camera poses by directly aligning learned
features~\cite{sarlin2021cvpr}, and recent approaches extend this idea to
differentiable Gaussian Splatting based scene representations
\cite{cheng2025icra,pietrantoni2025cvpr}.
We apply featuremetric optimization only to cross-session poses that have
passed geometric verification and sequence-level consistency checks.
The refined poses then serve as sparse constraints for complete trajectory
registration rather than as independent localization outputs.

%% BRIEFLY SUMMARIZE OWN CONTRIBUTION AT THE END

%%%%%%%%%%%%%%%%%%%%%%%%%%%%%%%%%%%%%%%%%%%%%%%%%%%%%%%%%%%%%%%%%%%%%%%%%%%%%%%%
% \section{Our Approach to \dots}
% \label{sec:main}
% %% Describe your approach. It is okay to divide the main section
% %%  into a few subsections (e.g., 2-4 subsections).

\section[
  Our Approach to Visual Multi-Session Registration
]{%
  Our Approach to\\
  Visual Multi-Session Registration%
}\label{sec:method}

\begin{figure*}[t]
  \resizebox{\textwidth}{!}{%
    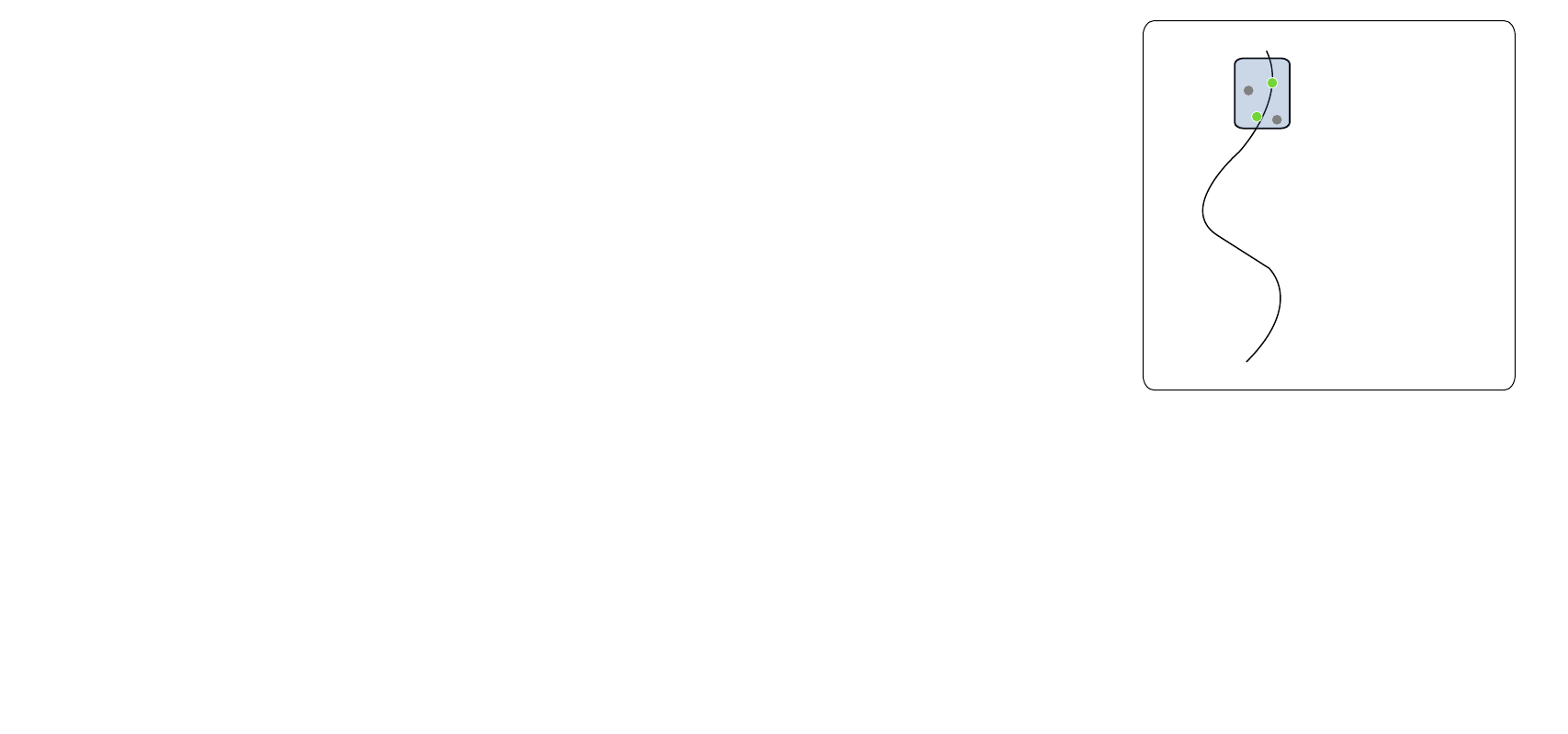}
\caption{%
  Overview of our visual multi-session registration pipeline.
  We first select distinctive cross-session retrievals and establish local feature correspondences between query and reference observations.
  Generalized PnP with RANSAC uses the reference-map geometry to reject correspondences that do not support a geometrically consistent multi-camera pose, producing locally verified absolute pose hypotheses.
  Since a locally plausible pose may still correspond to an incorrect location, sequence consensus evaluates the hypotheses jointly with the within-session query trajectory and retains only spatially distributed hypotheses that support a common query-to-reference similarity alignment.
  The resulting sparse absolute anchors are refined by featuremetric alignment against the reference Gaussian feature field and finally combined with within-session relative-pose constraints and consensus-aligned priors in a pose-graph optimization to recover the complete query trajectory.
}

  \label{fig:methodology}
\end{figure*}

We address registration of a visual query session to a fixed reference map of
the same environment built at a different time.
In challenging environments such as construction sites, substantial geometric
and appearance changes can limit reliable cross-session correspondences to only
a subset of query observations.
Our goal is to recover the complete query trajectory in the reference map frame
without directly localizing every query observation.

Our approach consists of five stages, illustrated in \figref{fig:methodology}.
First, we identify reliable cross-session retrieval hypotheses using visual
place recognition and distinctiveness filtering~(\secref{sec:retrieval}).
Second, we establish local feature correspondences and estimate absolute pose
hypotheses using reference-map geometry and generalized PnP~(\secref{sec:pnp}).
Third, we use the query trajectory to retain hypotheses supporting a common
query-to-reference alignment, forming sparse absolute anchors~(\secref{sec:consensus}).
Fourth, we refine these anchors through featuremetric alignment against the
reference feature field~(\secref{sec:featuremetric}).
Finally, we combine the refined anchors with within-session relative-pose
constraints in a robust pose-graph optimization (PGO) to recover the complete query trajectory
~(\secref{sec:pgo}).
\subsection{Preliminaries and Problem Formulation}
\label{sec:problem}

We consider a fixed reference map and a query session acquired in the
same environment at different times.
Both comprise synchronized multi-camera observations from a calibrated system
with $N_v$ cameras.
We index reference observations by time index $t$ and query observations by $k$, with
$N_q$ query observations in total.
Each observation has one body pose and one image from each camera~$v$.
We denote reference and query images by $I_{t,v}^M$ and $I_{k,v}^Q$,
respectively.

Coordinate frames are written as $\mathcal{F}_A$, and
$\m{T}_{AB} \in SE(3)$ maps coordinates from $\mathcal{F}_B$ to
$\mathcal{F}_A$.
Let $\mathcal{F}_M$ be the reference map coordinate frame and
$\mathcal{F}_Q$ the arbitrary query coordinate frame.
The body frame at observation $i$ is $\mathcal{F}_{B_i}$, and the frame of
camera $v$ at that observation is $\mathcal{F}_{C_{i,v}}$.
The fixed camera-to-body extrinsic calibration satisfies
$\m{T}_{B_iC_{i,v}}=\m{T}_{BC_v}\in SE(3)$ for all observations $i$.
The reference body poses $\m{T}_{MB_t}$ are known, and the corresponding
camera poses are
\begin{equation}
  \m{T}_{MC_{t,v}}
  =
  \m{T}_{MB_t}\m{T}_{B_tC_{t,v}}.
  \label{eq:reference_camera_pose}
\end{equation}

We represent the reference map using Gaussian
splatting~\cite{kerbl2023tog,huang2024siggraph} as a set of Gaussian primitives
encoding geometry, opacity, and appearance.
We additionally associate each Gaussian with a learned feature vector
$\v{f} \in \mathbb{R}^n$ and denote the resulting map by $\mathcal{G}$.
Given a camera pose $\m{T}_{MC}$, we render depth and feature maps as
\begin{equation}
  \m{D}
  =
  \mathcal{R}_D(\mathcal{G},\m{T}_{MC}),
  \quad
  \hat{\m{F}}
  =
  \mathcal{R}_F(\mathcal{G},\m{T}_{MC}),
  \label{eq:renderers}
\end{equation}
where $\mathcal{R}_D$ and $\mathcal{R}_F$ are the depth and feature
renderers~\cite{huang2024siggraph}, used for absolute pose estimation
and featuremetric pose refinement, respectively.

The query session provides body poses $\m{T}_{QB_k}$ in the arbitrary query
coordinate frame $\mathcal{F}_Q$.
We assume that their within-session relative geometry is consistent, while
their alignment with the reference map and, potentially, their metric scale
are unknown. Our objective is to estimate
\begin{equation}
  \mathcal{T}_Q^\star
  =
  \left\{
    \m{T}_{MB_k}^\star
  \right\}_{k=1}^{N_q},
  \label{eq:registration_objective}
\end{equation}
such that the complete query trajectory is expressed in the reference map
coordinate frame $\mathcal{F}_M$.

\subsection{Reliable Cross-Session Retrieval}
\label{sec:retrieval}
The first stage seeks to identify query observations for which a reliable
cross-session retrieval hypothesis can be established.
This is challenging in changing environments such as construction sites,
where repetitive structures and low-texture regions can produce perceptually
similar observations at physically distinct locations
\cite{torii2015tpami,sun2025isprs}.
Since an incorrect retrieval may lead to an erroneous absolute pose constraint,
we deliberately favor precision over retrieval density.

For every camera image $I_{k,v}^Q$ of query observation~$k$, a shared visual place recognition
encoder produces a global descriptor~$\v{d}_{k,v}^Q \in \mathbb{R}^m$.
We aggregate these descriptors into one descriptor for the complete observation,

\begin{equation}
  \v{d}_k^Q =
  \frac{\sum_{v=1}^{N_v} \v{d}_{k,v}^Q}
       {N_v}.   
  %  {\norm{\sum_{v=1}^{N_v} \v{d}_{k,v}^Q}}.
  \label{eq:observation_descriptor}
\end{equation}

We obtain reference descriptors $\v{d}_t^M$ analogously.
Given query observation~$k$ and reference observation~$t$, we measure their
visual similarity by
\begin{equation}
  s(k,t)
  =
  {\v{d}_k^Q}^{\mathsf T}\v{d}_t^M .
\end{equation}

The highest-scoring reference observation
\begin{equation}
  t^\star
  =
  \argmax_t s(k,t),
\end{equation}
defines the initial retrieval hypothesis.

The retrieval score $s(k,t)$ alone does not indicate whether the selected
location is visually distinctive from other parts of the environment.
One approach is to use the margin between the best and
second-ranked reference observations.
For sequentially acquired data, however, neighboring reference observations
often depict essentially the same place and consequently receive similarly
high similarity scores.
A small top-two margin therefore does not necessarily indicate genuine
spatial ambiguity.
We instead compare the best retrieval against the strongest spatially
distinct alternative.

Let $\v{p}_t$ denote the known position of reference observation~$t$ in the
reference map coordinate frame.
For an exclusion radius $r_{\mathrm{far}}$, we define the spatial margin as
\begin{equation}
  \begin{aligned}
    m_{\mathrm{far}}(k)
    &= s(k,t^\star)-\max_{t\in\mathcal{D}} s(k,t),\\
    \mathcal{D}
    &= \left\{t \mid
      \norm{\v{p}_t-\v{p}_{t^\star}}>r_{\mathrm{far}}
    \right\}.
  \end{aligned}
  \label{eq:far_margin}
\end{equation}

We accept retrievals with $m_{\mathrm{far}}(k)\geq\tau_{\mathrm{far}}$.
The exclusion radius prevents nearby observations of the same place from
competing, so a large margin indicates preference over visually similar
alternatives from spatially distinct map regions.

For each accepted query-reference pair, we use the per-camera
descriptors to identify views likely to depict the same scene content.
We compute pairwise descriptor similarities between query and reference
views and select the one-to-one assignment maximizing total similarity
using the Hungarian algorithm~\cite{kuhn1955naval}.
We then perform local feature matching only between the selected view pairs.
We lift the resulting 2D-2D correspondences to 3D using reference map
geometry and verify them with generalized PnP.

\subsection{Geometric Verification with Generalized PnP}
\label{sec:pnp}

The preceding stage provides tentative 2D--2D correspondences that need not be
geometrically consistent.
We therefore lift the reference matches to 3D using the reference map geometry
and estimate one shared multi-camera body pose through generalized PnP with
RANSAC.

Using the known reference camera pose $\m{T}_{MC_{t,v_r}}$, we render its depth
and back-project the reference pixel to obtain the 3D point
$\v{X}_n$ in the reference map coordinate frame:
\begin{equation}
  \begin{aligned}
    d_n
    &=
    \mathcal{R}_D(\mathcal{G},\m{T}_{MC_{t,v_r}})[\v{u}_n^M],\\
    \v{X}_n
    &=
    \m{T}_{MC_{t,v_r}}\,
    \pi_{v_r}^{-1}\left(\v{u}_n^M,d_n\right).
  \end{aligned}
  \label{eq:reference_backprojection}
\end{equation}

Here, $[\v{u}]$ denotes indexing the rendered depth map at pixel $\v{u}$.
Since rays from different cameras do not share a common optical center, let
$\v{o}_{v_q}$ and $\v{b}_n$ denote the camera center and corresponding unit bearing
vector expressed in $\mathcal{F}_{B_k}$.

We seek the body pose
$\m{T}_{MB_k}=
[\m{R}_{MB_k}\mid\v{t}_{MB_k}]$ obeying:
\begin{equation}
  \v{X}_n
  =
  \m{R}_{MB_k}
  \left(
    \v{o}_{v_q}
    +
    \lambda_n \v{b}_n
  \right)
  +
  \v{t}_{MB_k},
  \label{eq:gpnp}
\end{equation}
where $\lambda_n$ is the unknown ray depth.

We solve the generalized PnP problem jointly over all views using
RANSAC~\cite{kneip2013icra}.
RANSAC rejects inconsistent matches and returns an absolute pose hypothesis
$\m{T}_{MB_k}^{\mathrm{PnP}}$ from the inliers.
These locally verified hypotheses are then evaluated for consistency with the
query trajectory.
\subsection{Sequence Consensus}
\label{sec:consensus}

Generalized PnP locally verifies each absolute pose hypothesis
$\m{T}_{MB_k}^{\mathrm{PnP}}$, but a geometrically plausible pose may still
correspond to an incorrect place.
We therefore use the consistency of the query trajectory to identify hypotheses
that jointly support a common query-to-reference alignment.
For each candidate observation $k$, we have its pose $\m{T}_{QB_k}$ in the
query coordinate frame and its verified pose hypothesis
$\m{T}_{MB_k}^{\mathrm{PnP}}$ in the reference map coordinate frame.

We seek a similarity transformation with scale
\begin{equation}
  \m{S}_{MQ}
  =
  [s\m{R}_{MQ}\mid\v{t}_{MQ}],
  \qquad s>0,
\end{equation}
which aligns the query trajectory with the reference map while remaining
consistent with the absolute pose hypotheses.

This transformation predicts the reference-frame pose of query observation
$k$ as
\begin{equation}
  \tilde{\m{T}}_{MB_k}
  =
  \m{S}_{MQ}\m{T}_{QB_k}.
  \label{eq:consensus_prediction}
\end{equation}

A pose hypothesis $\m{T}_{MB_k}^{\mathrm{PnP}}$ supports the transformation
when its position and orientation agree with
$\tilde{\m{T}}_{MB_k}$.

To prevent dense neighboring observations from dominating the alignment, we
partition the query trajectory into spatial regions and favor support across
distinct regions.
We generate candidate transformations from spatially separated pose pairs.
Each proposal estimates rotation from the corresponding query-to-reference
orientations, and translation and, when required, scale from the associated
body centers.
We evaluate each proposal against all pose hypotheses using translation and
rotation residuals.

We select the proposal with the strongest spatially distributed support and
refit $\m{S}_{MQ}$ using its inliers.
During refitting, each spatial region receives equal total weight, preventing
densely sampled portions from dominating the estimate.
We recompute support and require the alignment to remain stable as supported
regions are removed.
If no sufficiently distributed and stable consensus exists, we reject the
session rather than forcing a global registration.

The resulting $\m{S}_{MQ}$ defines the consensus alignment between the query
trajectory and reference map.
Its inliers form the absolute anchor set $\mathcal{A}$ passed to featuremetric
refinement.
For each $k\in\mathcal{A}$, $\m{T}_{MB_k}^{\mathrm{PnP}}$ provides the initial anchor pose.

\begin{table*}[t]
  \centering
  \caption{%
    Aggregate multi-session registration performance over 15 cross-session
    experiments.
    Translation and rotation RMSE are computed only for poses with translation
    error below $5\,\mathrm{m}$, whereas recall is evaluated over all query
    observations.
    All metrics are micro-averaged over query observations.
  }
  \label{tab:main_results}
  \small
  \setlength{\tabcolsep}{3.5pt}
  \begin{tabular}{@{}lccccc@{}}
    \toprule
    Method
    & t-RMSE$_{<5\,\mathrm{m}}$ [m] $\downarrow$
    & r-RMSE$_{<5\,\mathrm{m}}$ [$^\circ$] $\downarrow$
    & $R@0.5 \mathrm{m}$ $\uparrow$
    & $R@1 \mathrm{m}$ $\uparrow$
    & $R@5 \mathrm{m}$ $\uparrow$ \\
    \midrule

    FPFH~\cite{rusu2009icra} + ICP~\cite{besl1992tpami}
    & 1.793
    & 36.51
    & 11.22\%
    & 34.59\%
    & 60.95\% \\

    Register Any Point~\cite{pan2026eccv}
    & 2.314
    & 69.50
    & 1.34\%
    & 6.69\%
    & 19.38\% \\

    ColabSfM~\cite{edstedt2025cvpr}
    & 2.254
    & 42.51
    & 5.46\%
    & 18.03\%
    & 46.47\% \\

    HLoc~\cite{sarlin2019cvpr}
    & 0.944
    & 15.21
    & 45.5\%
    & 65.3\%
    & 70.7\% \\

    \textbf{Ours}
    & \textbf{0.600}
    & \textbf{1.204}
    & \textbf{62.5\%}
    & \textbf{94.4\%}
    & \textbf{100.0\%} \\

    \bottomrule
  \end{tabular}
\end{table*}

\subsection{Featuremetric Pose Refinement}
\label{sec:featuremetric}
Sequence consensus provides sparse absolute anchors supported by local geometric
verification and consistency with the query trajectory.
Small residual pose errors can still cause noticeable image-to-map
misalignment.
We therefore refine each absolute anchor by aligning the query feature maps with the
reference feature field.
Since we refine only anchors retained by sequence consensus, we assume their
views contain sufficient persistent scene content for reliable alignment.
The learned features additionally provide robustness to appearance and texture
changes between sessions.

Let $\v{\xi}\in\RR^6$ denote a local pose update, with
$(\cdot)^\wedge:\RR^6\rightarrow\mathfrak{se}(3)$ mapping it to the Lie algebra
of $SE(3)$.
We parameterize the refined body pose as
\begin{equation}
  \m{T}_{MB_k}(\v{\xi})
  =
  \operatorname{Exp}(\v{\xi}^{\wedge})
  \m{T}_{MB_k}^{\mathrm{PnP}}.
  \label{eq:pose_update}
\end{equation}

For camera $v$, the pose and rendered feature map are
\begin{equation}
  \begin{aligned}
    \m{T}_{MC_{k,v}}(\v{\xi})
    &= \m{T}_{MB_k}(\v{\xi})\m{T}_{B_kC_{k,v}},\\
    \hat{\m{F}}_{k,v}(\v{\xi})
    &= \mathcal{R}_F(\mathcal{G},\m{T}_{MC_{k,v}}(\v{\xi})).
  \end{aligned}
  \label{eq:feature_render}
\end{equation}

Let $\Phi$ denote the feature encoder used to construct the reference feature
field and extract query features
$\m{F}_{k,v}^Q=\Phi(I_{k,v}^Q)$.
We optimize a shared body pose across cameras by minimizing the
query-rendered feature discrepancy:
\begin{equation}
  \v{\xi}^{\star}
  =
  \argmin_{\v{\xi}}
  \sum_{v=1}^{N_v}
  \frac{1}{|\Omega_v|}
  \sum_{\v{u}\in\Omega_v}
  \norm{
    \m{F}_{k,v}^Q[\v{u}]
    -
    \hat{\m{F}}_{k,v}(\v{\xi})[\v{u}]
  }_1 .
  \label{eq:featuremetric}
\end{equation}

Here, $\Omega_v$ denotes the valid rendered pixels for camera $v$.
Since the Gaussian renderer is differentiable with respect to pose, we
minimize~\eqref{eq:featuremetric} with $\mathcal{G}$ fixed, optimizing only the
six-degree-of-freedom body pose.
The resulting $\m{T}_{MB_k}(\v{\xi}^{\star})$ defines the refined anchor for
$k\in\mathcal{A}$ and is used in the subsequent pose-graph optimization.

\subsection{Pose-Graph Optimization}
\label{sec:pgo}

The preceding stages provide refined absolute anchors $\m{T}_{MB_k}(\v{\xi}^{\star})$ and the consensus
alignment $\m{S}_{MQ}$.
We initialize the query poses with the consensus-aligned poses
$\tilde{\m{T}}_{MB_k}$ from~\eqref{eq:consensus_prediction} and refine them
using PGO.

Let $\mathcal{E}$ denote the available within-session relative-pose constraints,
with measurements
\begin{equation}
  \m{Z}_{kl}
  =
  \m{T}_{QB_k}^{-1}\m{T}_{QB_l},
  \qquad (k,l)\in\mathcal{E}.
  \label{eq:relative_measurement}
\end{equation}

For trajectories with unknown metric scale, we apply the scale estimated during
sequence consensus to the relative-pose translations and keep it fixed during
optimization.

For each anchor $k\in\mathcal{A}$, let
$\m{A}_k=\m{T}_{MB_k}(\v{\xi}^{\star})$ denote its refined absolute pose.
We optimize the reference-frame body poses $\{\m{T}_{MB_k}\}$ using
relative-pose measurements $\m{Z}_{kl}$, refined anchors $\m{A}_k$, and
consensus-aligned pose priors $\tilde{\m{T}}_{MB_k}$:
\begin{equation}
\begin{split}
  \left\{\m{T}_{MB_k}^{\star}\right\}
  =
  \argmin_{\{\m{T}_{MB_k}\}}
  &\sum_{k\in\mathcal{A}}
  \rho
  \left(
    \norm{
      \operatorname{Log}
      \left(
        \m{A}_k^{-1}\m{T}_{MB_k}
      \right)
    }_{\boldsymbol{\Lambda}_A}^{2}
  \right)
  \\
  &\hspace{-23pt}+\,
  \sum_{(k,l)\in\mathcal{E}}
  \rho
  \left(
    \norm{
      \operatorname{Log}
      \left(
        \m{Z}_{kl}^{-1}
        \m{T}_{MB_k}^{-1}
        \m{T}_{MB_l}
      \right)
    }_{\boldsymbol{\Lambda}_R}^{2}
  \right)
  \\
  &\hspace{-23pt}+\,
  \sum_k
  \rho
  \left(
    \norm{
      \operatorname{Log}
      \left(
        \tilde{\m{T}}_{MB_k}^{-1}\m{T}_{MB_k}
      \right)
    }_{\boldsymbol{\Lambda}_P}^{2}
  \right).
\end{split}
\raisetag{20pt}
\label{eq:pgo}
\end{equation}

Here, $\rho$ is a robust loss, while the information matrices
$\boldsymbol{\Lambda}_A$, $\boldsymbol{\Lambda}_R$, and
$\boldsymbol{\Lambda}_P$ weight the anchor, relative-pose, and prior residuals.
The anchor and relative-pose factors constrain the trajectory to the refined
poses while preserving local motion, and the priors prevent poorly constrained
regions from drifting from the consensus alignment.
The optimized poses define $\mathcal{T}_Q^\star$
from~\eqref{eq:registration_objective}.
\subsection{Implementation Details}
We use MegaLoc~\cite{berton2025cvprws} for visual place recognition, LoMa-B~\cite{nordstrom2026eccv} for local
feature matching, and DINOv3 ViT-S/16~\cite{simeoni2025arxiv} for featuremetric refinement.
MegaLoc descriptors are extracted from $322\times322$ images.
For the spatial far-margin, we use $r_{\mathrm{far}}=10\,\mathrm{m}$ and accept
retrievals with $m_{\mathrm{far}}\geq0.01$.
For each accepted retrieval, up to eight one-to-one query-reference view pairs
are matched with LoMa-B using at most 4096 keypoints per image.
Generalized PnP with RANSAC uses a $12$\,px reprojection threshold, confidence
$0.999$, at most 1000 iterations, and requires at least 20 inliers.

For sequence consensus, spatial cells have width $5\%$ of the query trajectory
bounding-box diagonal.
Translation and rotation inlier thresholds are $15\%$ in normalized trajectory
coordinates and $15^\circ$, respectively, and a valid consensus requires at
least six anchors across three spatial cells.
Featuremetric refinement optimizes one shared $SE(3)$ body-pose update over all
valid camera views using Adam for 32 steps with learning rate $10^{-3}$.
Pose corrections are limited to $0.03$ in normalized translation and
$3^\circ$ in rotation.

The final pose graph uses a soft-$L_1$ loss.
Translation standard deviations for relative-pose, absolute-anchor, and
trajectory-prior factors are $0.005$, $0.03$, and $0.05$, with corresponding
rotation standard deviations of $0.5^\circ$, $2^\circ$, and $3^\circ$.
Scale recovered by sequence consensus is fixed before PGO.
%%%%%%%%%%%%%%%%%%%%%%%%%%%%%%%%%%%%%%%%%%%%%%%%%%%%%%%%%%%%%%%%%%%%%%%%%%%%%%%%

\section{Experimental Evaluation}
\label{sec:exp}

We evaluate our approach for registering visual sessions acquired at different
times to an earlier reference map under substantial scene change and perceptual
ambiguity.
Specifically, we evaluate three claims:
(i) our approach provides more accurate and complete multi-session registration
than the baselines;
(ii) conservative cross-session filtering and anchor selection improve the
reliability of absolute pose constraints under scene change and perceptual
aliasing; and
(iii) featuremetric refinement improves the fine image-to-map alignment of
verified absolute anchors.

\subsection{Experimental Setup}

We use the HILTI26~\cite{centanni2026arxiv} construction-site dataset, where
data were collected by human operators and ground truth was generated using
LiDAR-Inertial SLAM.
We construct 15 cross-session registration experiments using seven reference sessions.
Unlike HILTI23~\cite{nair2024ral}, whose multi-session recordings span only
several days, HILTI26 follows an active construction project over eight
months and therefore captures substantially larger structural changes between sessions.
We exclude floor~5 because it has only one recording, and UG2, floor~7, and the
second runs of floor~2 and UG1 due to SfM failures to generate even a single
consistent map.
For each remaining environment, we use the earliest recording as the reference
session and register all later recordings against it, evaluating registration
against an earlier map as the construction site evolves.

For each reference session, we reconstruct the scene using SfM and train a
3D Gaussian Splatting map~\cite{kerbl2023tog} from the resulting images
and poses.
Within-session relative-pose constraints for each query session are obtained
from an independent SfM reconstruction.
Only its relative geometry is provided; its alignment and metric scale with
respect to the reference map must be recovered during registration.
All experiments run on an NVIDIA RTX A5000 GPU, Intel Xeon W-1290P CPU, and
64\,GiB of RAM.

%%%%%%%%%%%%%%%%%%%%%%%%%%%%%%%%%%%%%%%%%%%%%%%%%%%%%%%%%%%%%%%%%%%%%%%%%%%%%%%%
\begin{figure}[t]
  \centering
  \includegraphics[width=\linewidth]{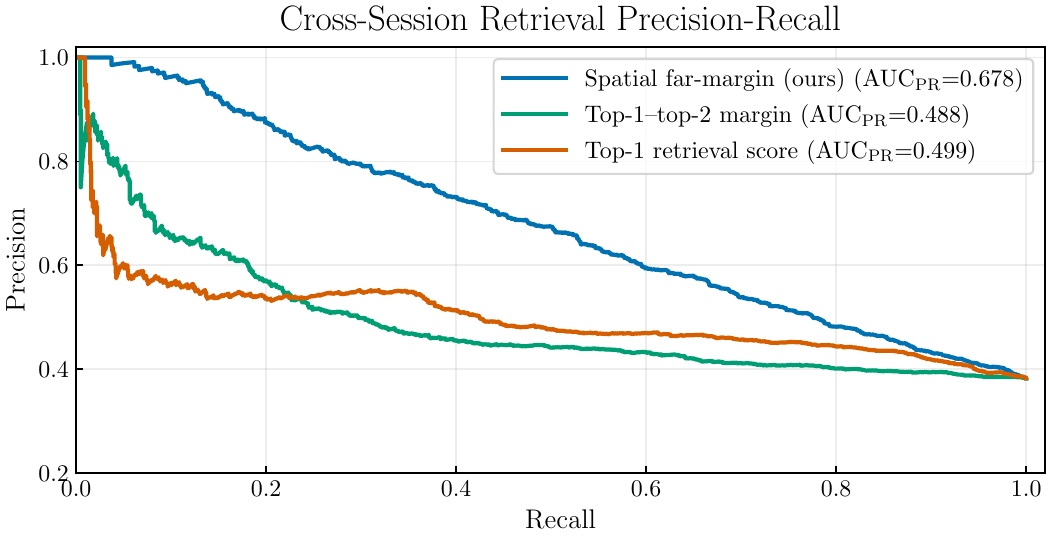}
  \caption{%
  Precision--recall trade-off of VPR confidence criteria.
  Retrievals are considered correct if the retrieved reference observation lies within
  $2\,\mathrm{m}$ of the ground-truth position.
  Our spatial far-margin achieves substantially higher precision across a broad
  range of recall than the top-1 retrieval score and the top-1--top-2 margin.
  }
  \label{fig:vpr_precision_coverage}
\end{figure}

\subsection{Multi-Session Registration}
The first experiment evaluates the accuracy and completeness of our
multi-session registration and compares its performance against several
baselines. Its results directly support our first claim that our approach
provides more accurate and complete registration than existing alternatives.

We compare against HLoc~\cite{sarlin2019cvpr}, using the same MegaLoc retrieval,
LoMa-B matching, and generalized multi-camera PnP configuration as our
localization front end to ensure a fair comparison. 
HLoc localizes each query observation independently against the reference map
and does not exploit within-session relative-pose constraints.

As a complementary geometry-based baseline, we independently reconstruct the
reference and query sessions and register their sparse SfM point clouds using
FPFH-based~\cite{rusu2009icra} RANSAC followed by ICP~\cite{besl1992tpami}.
Since their relative scale is unknown, we provide this baseline with the
ground-truth scale, removing scale estimation as a source of failure. 
We additionally evaluate recent learned registration methods,
ColabSfM~\cite{edstedt2025cvpr} and Register Any Point~\cite{pan2026eccv},
on the same cross-session SfM reconstructions.

\tabref{tab:main_results} reports the aggregate registration performance.
Our approach achieves the highest recall at all thresholds, with
$R@0.5\,\mathrm{m}=62.5\%$, $R@1\,\mathrm{m}=94.4\%$, and
$R@5\,\mathrm{m}=100.0\%$.
Among poses with translation error below $5\,\mathrm{m}$, it also achieves the
lowest translation and rotation RMSE, with $0.600\,\mathrm{m}$ and
$1.204^\circ$, respectively.

HLoc performs well for observations that can be localized independently, but
suffers from catastrophic failures when reliable cross-session correspondences
are unavailable or ambiguous.
Direct registration of independently reconstructed SfM point clouds is
considerably less reliable.
Register Any Point is designed as a general-purpose point-cloud registration
method rather than specifically for sparse SfM geometry or temporally changing
scenes~\cite{pan2026eccv}, while ColabSfM is trained on trajectories observing
the same underlying scene~\cite{edstedt2025cvpr}.
Both therefore face a substantial domain shift under the structural changes
present in our setting.
FPFH-based RANSAC followed by ICP is more robust than the learned registration
methods, but still remains well below our image-based registration approach.

These results support our first claim: sparse, reliable cross-session constraints
combined with within-session trajectory information provide more complete and
accurate registration than the evaluated alternatives.

\begin{figure*}[t]
  \centering
  \resizebox{0.90\linewidth}{!}{%
    % \import{pics/}{qualitative_fig.pdf}}
    \includegraphics{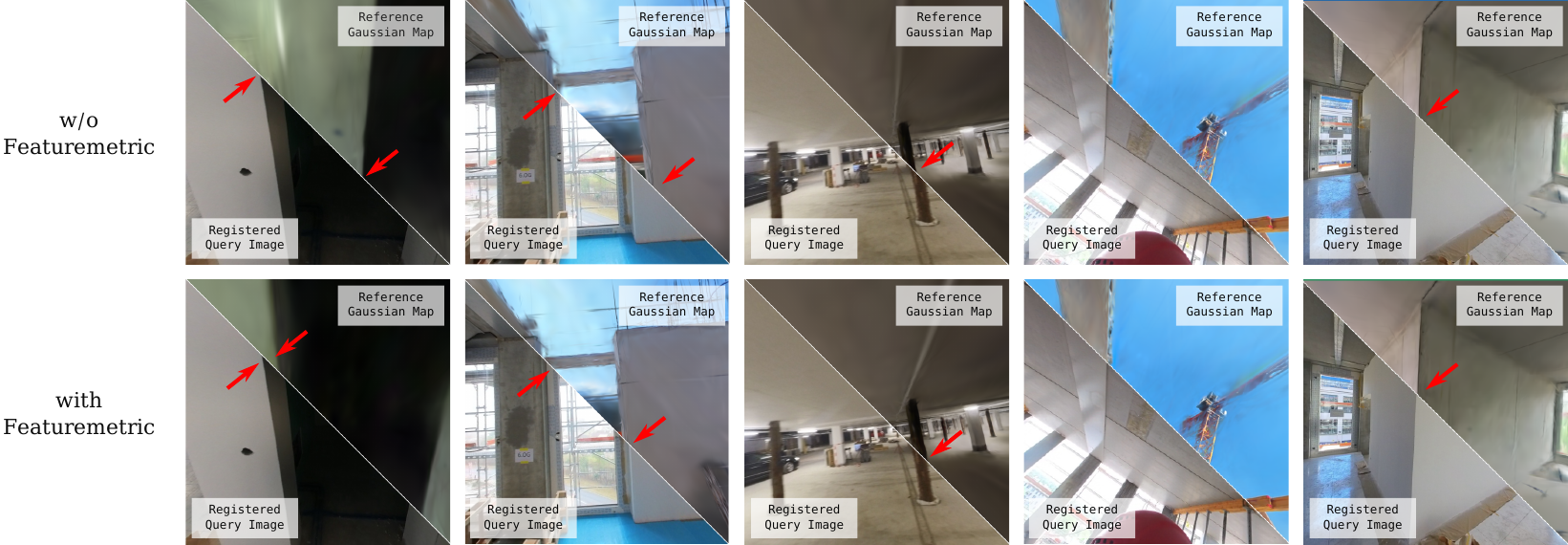}}
\caption{%
  Qualitative effect of featuremetric pose refinement.
  Top row shows alignment after sequence consensus and PGO. Bottom row additionally uses featuremetric refinement, which corrects residual misalignment when present while preserving already well-aligned poses.
}
  \label{fig:featuremetric_alignment}
\end{figure*}

%%%%%%%%%%%%%%%%%%%%%%%%%%%%%%%%%%%%%%%%%%%%%%%%%%%%%%%%%%%%%%%%%%%%%%%%%%%%%%%%
\subsection{Reliability of Cross-Session Constraints}

The second experiment evaluates the reliability of the cross-session pose
constraints produced by our retrieval and geometric verification pipeline.
We first isolate the VPR distinctiveness criterion, then test whether
sequence consensus rejects locally plausible but globally inconsistent
hypotheses.

\begin{table}[t]
  \centering
  \caption{%
    Ablation of sequence consensus, pose-graph optimization, and featuremetric
    refinement.
    Unlike \tabref{tab:main_results}, RMSE is computed without an error
    gate.
  }
  \label{tab:registration_ablation}
  \small
  \setlength{\tabcolsep}{2.5pt}
  \begin{tabular}{@{}lccc@{}}
    \toprule
    Configuration
    & \shortstack{t-RMSE\\{[m]}}
    & \shortstack{r-RMSE\\{[$^\circ$]}}
    & $R@0.5$ \\
    \midrule

    Raw PnP hypotheses
    & 12.398
    & 68.144
    & 25.7\% \\

    PnP + PGO
    & 0.862
    & 2.297
    & 45.2\% \\

    PnP + Consensus alignment
    & 0.681
    & 1.334
    & 49.9\% \\

    PnP + Cons. + PGO
    & 0.608
    & 1.218
    & 60.7\% \\

    PnP + Cons. + Featuremetric + PGO
    & \textbf{0.600}
    & \textbf{1.204}
    & \textbf{62.5\%} \\

    \bottomrule
  \end{tabular}
\end{table}

\paragraph{Visual Distinctiveness}
We compare the proposed spatial far-margin against the top-1 retrieval score
and conventional top-1--top-2 margin.
\figref{fig:vpr_precision_coverage} shows the resulting precision-recall
trade-off, where a retrieval is correct if the retrieved reference observation
lies within $\tau_{\mathrm{VPR}}=2\,\mathrm{m}$ of the ground-truth query position.

The spatial far-margin achieves higher precision over a broad range of recall.
Unlike the conventional top-two margin, it does not penalize neighboring
reference observations depicting the same physical location.
Instead, it measures whether the selected location remains distinctive from
spatially separated alternatives.
This matters in repetitive construction environments, where perceptually similar
observations may originate from distant site regions.

\paragraph{Sequence Consensus}

We then evaluate whether sequence consensus rejects absolute pose hypotheses
that are locally plausible but inconsistent with the query trajectory.
\tabref{tab:registration_ablation} compares raw generalized PnP hypotheses,
PGO without sequence consensus, consensus alignment, and the subsequent
refinement stages.
Raw PnP hypotheses are available only for directly localized observations,
whereas the remaining configurations estimate the complete query trajectory.
Recall is evaluated over all query observations, and RMSE is computed without
the $5\,\mathrm{m}$ gate used in \tabref{tab:main_results}.

Raw generalized-PnP hypotheses contain severe outliers, with a translation RMSE
of $12.4\,\mathrm{m}$, rotation RMSE of $68.1^\circ$, and
$R@0.5\,\mathrm{m}=25.7\%$.
Directly combining these hypotheses with within-session relative-pose constraints
using PGO substantially reduces the error, but remains less reliable, with
$0.862\,\mathrm{m}$ translation RMSE, $2.3^\circ$ rotation RMSE, and
$R@0.5\,\mathrm{m}=45.2\%$.
Thus, PGO alone cannot fully compensate for incorrect absolute constraints.

Sequence consensus further improves registration reliability.
Consensus alignment alone reduces translation and rotation RMSE to
$0.681\,\mathrm{m}$ and $1.3^\circ$, while increasing
$R@0.5\,\mathrm{m}$ to $49.9\%$.
Adding PGO further improves $R@0.5\,\mathrm{m}$ to $60.7\%$ and
translation RMSE to $0.608\,\mathrm{m}$.

These results support our second claim that filtering absolute pose hypotheses
before trajectory optimization produces more reliable cross-session constraints
than jointly optimizing all locally verified hypotheses.
%%%%%%%%%%%%%%%%%%%%%%%%%%%%%%%%%%%%%%%%%%%%%%%%%%%%%%%%%%%%%%%%%%%%%%%%%%%%%%%%

\subsection{Featuremetric Pose Refinement}

Finally, the last experiment evaluates whether
featuremetric refinement improves the fine image-to-map alignment of the
verified absolute anchors.
Unlike the preceding filtering stages, featuremetric optimization is not
intended to recover from incorrect place hypotheses or large pose errors.
Instead, it operates only on the absolute anchors retained by sequence consensus
and targets the remaining small image-to-map misalignments.

As shown in \tabref{tab:registration_ablation}, adding featuremetric
refinement before PGO reduces the translation RMSE from
$0.608\,\mathrm{m}$ to $0.600\,\mathrm{m}$ and also slightly reduces the
rotation RMSE.
Registration recall at the stricter threshold, $R@0.5\,\mathrm{m}$, increases
from $60.7\%$ to $62.5\%$.
This modest trajectory-level improvement is expected because refinement is
applied only to already verified sparse anchors, whose pose corrections are
subsequently propagated through the trajectory by PGO.

The effect of featuremetric refinement is more apparent at the image level.
\figref{fig:featuremetric_alignment} shows representative absolute anchors
before and after optimization.
The generalized-PnP poses already provide geometrically valid
initializations, while featuremetric refinement improves the alignment of
persistent structures between the real query observations and the reference
Gaussian rendering despite scene changes. This is a clear support of our third claim

% % Optional robustness experiment if space permits.
% \paragraph{Effect of Temporal Separation.}
% Finally, we examine how registration performance changes as the temporal
% separation between the reference and query sessions increases.
% Using session pairs acquired at different stages of construction, we relate the
% time interval between acquisitions to registration accuracy and recall.
% This experiment characterizes the degradation caused by increasing scene
% evolution rather than introducing an additional component of the method.

% \begin{figure}[t]
%   \centering
%   \fbox{%
%     \rule{0pt}{0.18\textwidth}%
%     \rule{0.45\textwidth}{0pt}%
%   }
%   \caption{%
%     Registration performance as a function of the temporal separation between
%     reference and query sessions.
%   }
%   \label{fig:temporal_gap}
% \end{figure}

% In summary, the experiments evaluate the complete registration performance,
% the reliability of the selected cross-session anchors, and the contribution of
% featuremetric refinement.
% Together, they test the three claims stated in Sec.~\ref{sec:intro}.

%%%%%%%%%%%%%%%%%%%%%%%%%%%%%%%%%%%%%%%%%%%%%%%%%%%%%%%%%%%%%%%%%%%%%%%%%%%%%%%%
\section{Conclusion}
\label{sec:conclusion}

In this paper, we presented a novel, complete visual multi-session registration
pipeline designed for challenging and changing environments such as construction sites,
without requiring reliable cross-session correspondences along the entire
query trajectory.
Our method identifies a sparse set of strongly supported absolute anchors
using visual distinctiveness, geometric verification, and sequence-level
consistency, refines these anchors against a reference Gaussian feature field,
and propagates their information through the query trajectory using
within-session relative-pose constraints.

Our experimental evaluation on evolving construction environments shows that
this strategy provides more complete and accurate trajectory registration than
independent image localization and direct registration of independently
reconstructed geometry.
If no sufficiently distributed consensus is found, our method rejects the
session rather than forcing an incorrect global alignment.
The results further demonstrate the importance of conservative anchor
selection and show that featuremetric refinement improves the fine
image-to-map alignment of verified absolute anchors.

These properties make the proposed approach suitable for persistent mapping
and repeated robotic operation under substantial structural change.

%%%%%%%%%%%%%%%%%%%%%%%%%%%%%%%%%%%%%%%%%%%%%%%%%%%%%%%%%%%%%%%%%%%%%%%%%%%%%%%%
%% Future work: Use only if applicable -- but if so, use the following
%% sentence to start:
% Despite these encouraging results, there is further space for improvements. 
%% In general, I avoid explaining future work in 6-8 page conference papers...

%%%%%%%%%%%%%%%%%%%%%%%%%%%%%%%%%%%%%%%%%%%%%%%%%%%%%%%%%%%%%%%%%%%%%%%%%%%%%%%%
% Only if applicable
%\section*{Acknowledgments}
%We thank XXX for fruitful discussions and for \dots

\flushcolsend
\clearpage
\raggedcolsend
\bibliographystyle{plain_abbrv}

% All new citations should go to new.bib. The file glorified.bib should go
% be the one from the ipb server. After paper or related work has been
% written merge the entries from new.bib to glorified.bib ON THE SERVER,
% replace the glorified.bib in this repository and empty the new.bib
\bibliography{glorified,new}

\begin{thebibliography}{10}

\bibitem{alibey2023wacv}
A.~Ali-bey, B.~Chaib-draa, and P.~Gigu{\`e}re.
\newblock {MixVPR: Feature Mixing for Visual Place Recognition}.
\newblock In {\em Proc.~of the IEEE Winter Conf.~on Applications of Computer
  Vision (WACV)}, 2023.

\bibitem{arandjelovic2016cvpr}
R.~Arandjelovic, P.~Gronat, A.~Torii, T.~Pajdla, and J.~Sivic.
\newblock Netvlad: Cnn architecture for weakly supervised place recognition.
\newblock In {\em Proc.~of the IEEE Conf.~on Computer Vision and Pattern
  Recognition (CVPR)}, 2016.

\bibitem{berton2025cvprws}
G.~Berton and C.~Masone.
\newblock {MegaLoc: One Retrieval to Place Them All}.
\newblock In {\em Proc.~of the IEEE/CVF Conf. on Computer Vision and Pattern
  Recognition Workshops}, 2025.

\bibitem{berton2022cvpr}
G.~Berton, C.~Masone, and B.~Caputo.
\newblock {Rethinking Visual Geo-Localization for Large-Scale Applications}.
\newblock In {\em Proc.~of the IEEE/CVF Conf.~on Computer Vision and Pattern
  Recognition (CVPR)}, 2022.

\bibitem{berton2024ral}
G.~Berton, G.~Trivigno, B.~Caputo, and C.~Masone.
\newblock {JIST: Joint Image and Sequence Training for Sequential Visual Place
  Recognition}.
\newblock {\em IEEE Robotics and Automation Letters (RA-L)}, 9(2):1310--1317,
  2024.

\bibitem{besl1992tpami}
P.J. Besl and N.D. McKay.
\newblock {A Method for Registration of 3-D Shapes}.
\newblock {\em IEEE Trans.~on Pattern Analysis and Machine Intelligence
  (TPAMI)}, 14(2):239--256, 1992.

\bibitem{campos2021tro}
C.~Campos, R.~Elvira, J.J.G. Rodr{\'\i}guez, J.M. Montiel, and J.D. Tard{\'o}s.
\newblock Orb-slam3: An accurate open-source library for visual,
  visual--inertial, and multimap slam.
\newblock {\em IEEE Trans.~on Robotics (TRO)}, 37(6):1874--1890, 2021.

\bibitem{centanni2026arxiv}
S.~Centanni, Y.~Zhang, Y.~Tao, J.~Kindle, F.~Neuhaus, T.~Ko{\ss}, A.~Patel,
  M.~Helmberger, E.~Szyma{\'n}ska, T.~Gr{\"a}ber, and M.~Fallon.
\newblock {{Hilti}-{Trimble}-{Oxford} Dataset: 360 Visual-Inertial Benchmark
  with Floor Plan Priors for SLAM and Localization}.
\newblock {\em arXiv preprint}, arXiv:2607.06464, 2026.

\bibitem{cheng2025icra}
Y.~Cheng, J.~Jiao, Y.~Wang, and D.~Kanoulas.
\newblock {LoGS: Visual Localization via Gaussian Splatting with Fewer Training
  Images}.
\newblock In {\em Proc.~of the IEEE Intl.~Conf.~on Robotics \& Automation
  (ICRA)}, 2025.

\bibitem{edstedt2025cvpr}
J.~Edstedt, A.~Mateus, and A.~Jaenal.
\newblock Colabsfm: Collaborative structure-from-motion by point cloud
  registration.
\newblock In {\em Proc.~of the IEEE/CVF Conf.~on Computer Vision and Pattern
  Recognition (CVPR)}, 2025.

\bibitem{hausler2025ral}
S.~Hausler and P.~Moghadam.
\newblock {Pair-VPR: Place-Aware Pre-Training and Contrastive Pair
  Classification for Visual Place Recognition With Vision Transformers}.
\newblock {\em IEEE Robotics and Automation Letters (RA-L)}, 10(4):4013--4020,
  2025.

\bibitem{huang2024siggraph}
B.~Huang, Z.~Yu, A.~Chen, A.~Geiger, and S.~Gao.
\newblock {2D Gaussian Splatting for Geometrically Accurate Radiance Fields}.
\newblock In {\em Proc.~of the Intl.~Conf.~on Computer Graphics and Interactive
  Techniques (SIGGRAPH)}, 2024.

\bibitem{hutter2018fsr}
M.~Hutter, R.~Diethelm, S.~Bachmann, P.~Fankhauser, C.~Gehring, V.~Tsounis,
  A.~Lauber, F.~Guenther, M.~Bjelonic, L.~Isler, H.~Kolvenbach, K.~Meyer, and
  M.A. Hoepflinger.
\newblock {Towards a Generic Solution for Inspection of Industrial Sites}.
\newblock In {\em Field and Service Robotics}, 2018.

\bibitem{izquierdo2024cvpr}
S.~Izquierdo and J.~Civera.
\newblock {Optimal Transport Aggregation for Visual Place Recognition}.
\newblock In {\em Proc.~of the IEEE/CVF Conf.~on Computer Vision and Pattern
  Recognition (CVPR)}, 2024.

\bibitem{keetha2024ral}
N.V. Keetha, A.~Mishra, J.~Karhade, K.M. Jatavallabhula, S.~Scherer,
  M.~Krishna, and S.~Garg.
\newblock {AnyLoc: Towards Universal Visual Place Recognition}.
\newblock {\em IEEE Robotics and Automation Letters (RA-L)}, 9(2):1286--1293,
  2024.

\bibitem{kerbl2023tog}
B.~Kerbl, G.~Kopanas, T.~Leimk{\"u}hler, and G.~Drettakis.
\newblock {3D Gaussian Splatting for Real-Time Radiance Field Rendering}.
\newblock {\em ACM Trans.~on Graphics (TOG)}, 42(4):1--14, 2023.

\bibitem{kim2013tro}
A.~Kim and R.M. Eustice.
\newblock {Real-Time Visual SLAM for Autonomous Underwater Hull Inspection
  Using Visual Saliency}.
\newblock {\em IEEE Trans.~on Robotics (TRO)}, 29(3):719--733, 2013.

\bibitem{kneip2013icra}
L.~Kneip, P.~Furgale, and R.~Siegwart.
\newblock {Using multi-camera systems in robotics: Efficient solutions to the
  NPnP problem}.
\newblock In {\em Proc.~of the IEEE Intl.~Conf.~on Robotics \& Automation
  (ICRA)}, 2013.

\bibitem{kuhn1955naval}
H.W. Kuhn.
\newblock {The Hungarian Method for the Assignment Problem}.
\newblock {\em Naval Research Logistics Quarterly}, 2(1--2), 1955.

\bibitem{li2025ral}
H.~Li, G.~Peng, J.~Zhang, M.~Wen, Y.~Ma, and D.~Wang.
\newblock {CaseVPR: Correlation-Aware Sequential Embedding for
  Sequence-to-Frame Visual Place Recognition}.
\newblock {\em IEEE Robotics and Automation Letters (RA-L)}, 10(4):3430--3437,
  2025.

\bibitem{lim2025icra}
H.~Lim, D.~Kim, G.~Shin, J.~Shi, I.~Vizzo, H.~Myung, J.~Park, and L.~Carlone.
\newblock {KISS-Matcher: Fast and Robust Point Cloud Registration Revisited}.
\newblock In {\em Proc.~of the IEEE Intl.~Conf.~on Robotics \& Automation
  (ICRA)}, 2025.

\bibitem{nair2024ral}
A.D. Nair, J.~Kindle, P.~Levchev, and D.~Scaramuzza.
\newblock {Hilti SLAM Challenge 2023: Benchmarking Single + Multi-Session SLAM
  Across Sensor Constellations in Construction}.
\newblock {\em IEEE Robotics and Automation Letters (RA-L)}, 9(8):7286--7293,
  2024.

\bibitem{nordstrom2026eccv}
D.~Nordström, J.~Edstedt, G.~Bökman, J.~Astermark, A.~Heyden, V.~Larsson,
  M.~Wadenbäck, M.~Felsberg, and F.~Kahl.
\newblock Loma: Local feature matching revisited.
\newblock In {\em Proc.~of the Europ.~Conf.~on Computer Vision (ECCV)}, 2026.

\bibitem{pan2026eccv}
Y.~Pan, T.~Sun, L.~Zhu, L.~Nunes, I.~Armeni, J.~Behley, and C.~Stachniss.
\newblock {Register Any Point: Scaling 3D Point Cloud Registration by Flow
  Matching}.
\newblock In {\em Proc.~of the Europ.~Conf.~on Computer Vision (ECCV)}, 2026.

\bibitem{pietrantoni2025cvpr}
M.~Pietrantoni, G.~Csurka, and T.~Sattler.
\newblock {Gaussian Splatting Feature Fields for (Privacy-Preserving) Visual
  Localization}.
\newblock In {\em Proc.~of the IEEE/CVF Conf.~on Computer Vision and Pattern
  Recognition (CVPR)}, 2025.

\bibitem{rusu2009icra}
R.~Rusu, N.~Blodow, and M.~Beetz.
\newblock Fast point feature histograms (fpfh) for 3d registration.
\newblock In {\em Proc.~of the IEEE Intl.~Conf.~on Robotics \& Automation
  (ICRA)}, 2009.

\bibitem{sarlin2019cvpr}
P.E. Sarlin, C.~Cadena, R.~Siegwart, and M.~Dymczyk.
\newblock {From Coarse to Fine: Robust Hierarchical Localization at Large
  Scale}.
\newblock In {\em Proc.~of the IEEE/CVF Conf.~on Computer Vision and Pattern
  Recognition (CVPR)}, 2019.

\bibitem{sarlin2021cvpr}
P.E. Sarlin, A.~Unagar, M.~Larsson, H.~Germain, C.~Toft, V.~Larsson,
  M.~Pollefeys, V.~Lepetit, L.~Hammarstrand, F.~Kahl, and T.~Sattler.
\newblock {Back to the Feature: Learning Robust Camera Localization from Pixels
  to Pose}.
\newblock In {\em Proc.~of the IEEE/CVF Conf.~on Computer Vision and Pattern
  Recognition (CVPR)}, 2021.

\bibitem{sattler2018cvpr}
T.~Sattler, W.~Maddern, C.~Toft, A.~Torii, L.~Hammarstrand, E.~Stenborg,
  D.~Safari, M.~Okutomi, M.~Pollefeys, J.~Sivic, F.~Kahl, and T.~Pajdla.
\newblock {Benchmarking 6DOF Outdoor Visual Localization in Changing
  Conditions}.
\newblock In {\em Proc.~of the IEEE/CVF Conf.~on Computer Vision and Pattern
  Recognition (CVPR)}, 2018.

\bibitem{simeoni2025arxiv}
O.~Siméoni, H.V. Vo, M.~Seitzer, F.~Baldassarre, M.~Oquab, C.~Jose,
  V.~Khalidov, M.~Szafraniec, S.~Yi, M.~Ramamonjisoa, F.~Massa, D.~Haziza,
  L.~Wehrstedt, J.~Wang, T.~Darcet, T.~Moutakanni, L.~Sentana, C.~Roberts,
  A.~Vedaldi, J.~Tolan, J.~Brandt, C.~Couprie, J.~Mairal, H.~Jégou,
  P.~Labatut, and P.~Bojanowski.
\newblock Dinov3.
\newblock {\em Trans.~on Machine Learning Research (TMLR)}, 2026.

\bibitem{stenborg2020threedv}
E.~Stenborg, T.~Sattler, and L.~Hammarstrand.
\newblock {Using Image Sequences for Long-Term Visual Localization}.
\newblock In {\em Proc.~of the Intl.~Conf.~on 3D Vision (3DV)}, 2020.

\bibitem{sun2025isprs}
T.~Sun, Y.~Hao, S.~Huang, S.~Savarese, K.~Schindler, M.~Pollefeys, and
  I.~Armeni.
\newblock {Nothing Stands Still: A Spatiotemporal Benchmark on 3D Point Cloud
  Registration Under Large Geometric and Temporal Change}.
\newblock {\em ISPRS Journal of Photogrammetry and Remote Sensing},
  220:799--823, 2025.

\bibitem{torii2015tpami}
A.~Torii, J.~Sivic, M.~Okutomi, and T.~Pajdla.
\newblock Visual place recognition with repetitive structures.
\newblock {\em IEEE Trans.~on Pattern Analysis and Machine Intelligence
  (TPAMI)}, 37(11):2346--2359, 2015.

\bibitem{tuttas2014isprs}
S.~Tuttas, A.~Braun, A.~Borrmann, and U.~Stilla.
\newblock {Acquisition and Consecutive Registration of Photogrammetric Point
  Clouds for Construction Progress Monitoring Using a 4D BIM}.
\newblock In {\em ISPRS Archives of the Photogrammetry, Remote Sensing and
  Spatial Information Sciences}, 2014.

\bibitem{vysotska2019ral}
O.~Vysotska and C.~Stachniss.
\newblock {Effective Visual Place Recognition Using Multi-Sequence Maps}.
\newblock {\em IEEE Robotics and Automation Letters (RA-L)}, 4(2):1730--1736,
  2019.

\bibitem{vysotska2025icra}
O.~Vysotska, I.~Bogoslavskyi, M.~Hutter, and C.~Stachniss.
\newblock {Adaptive Thresholding for Sequence-Based Place Recognition}.
\newblock In {\em Proc.~of the IEEE Intl.~Conf.~on Robotics \& Automation
  (ICRA)}, 2025.

\bibitem{wang2020icra}
Y.~Wang, M.~Ramezani, and M.~Fallon.
\newblock {Actively Mapping Industrial Structures with Information Gain-Based
  Planning on a Quadruped Robot}.
\newblock In {\em Proc.~of the IEEE Intl.~Conf.~on Robotics \& Automation
  (ICRA)}, 2020.

\bibitem{yang2021tro}
H.~Yang, J.~Shi, and L.~Carlone.
\newblock {TEASER: Fast and Certifiable Point Cloud Registration}.
\newblock {\em IEEE Trans.~on Robotics (TRO)}, 37(2):314--333, 2021.

\end{thebibliography}

\end{document}